\documentclass[10pt,twocolumn,letterpaper]{article}

\usepackage[pagenumbers]{cvpr}        

\usepackage{amsmath}
\usepackage{booktabs}
\usepackage{graphicx}
\usepackage{array}

\definecolor{cvprblue}{rgb}{0.21,0.49,0.74}
\usepackage[pagebackref,breaklinks,colorlinks,allcolors=cvprblue]{hyperref}

\def\paperID{*****}
\def\confName{CVPR}
\def\confYear{2026}

\title{Multi-Stage VLM Pipeline for Zero-Shot Traffic Accident Understanding}

\author{Fumiya Tatematsu \quad Fumihiko Takahashi \\ GO Drive Inc.}

\begin{document}
\maketitle

\begin{figure*}[t]
\centering
\includegraphics[width=0.98\linewidth]{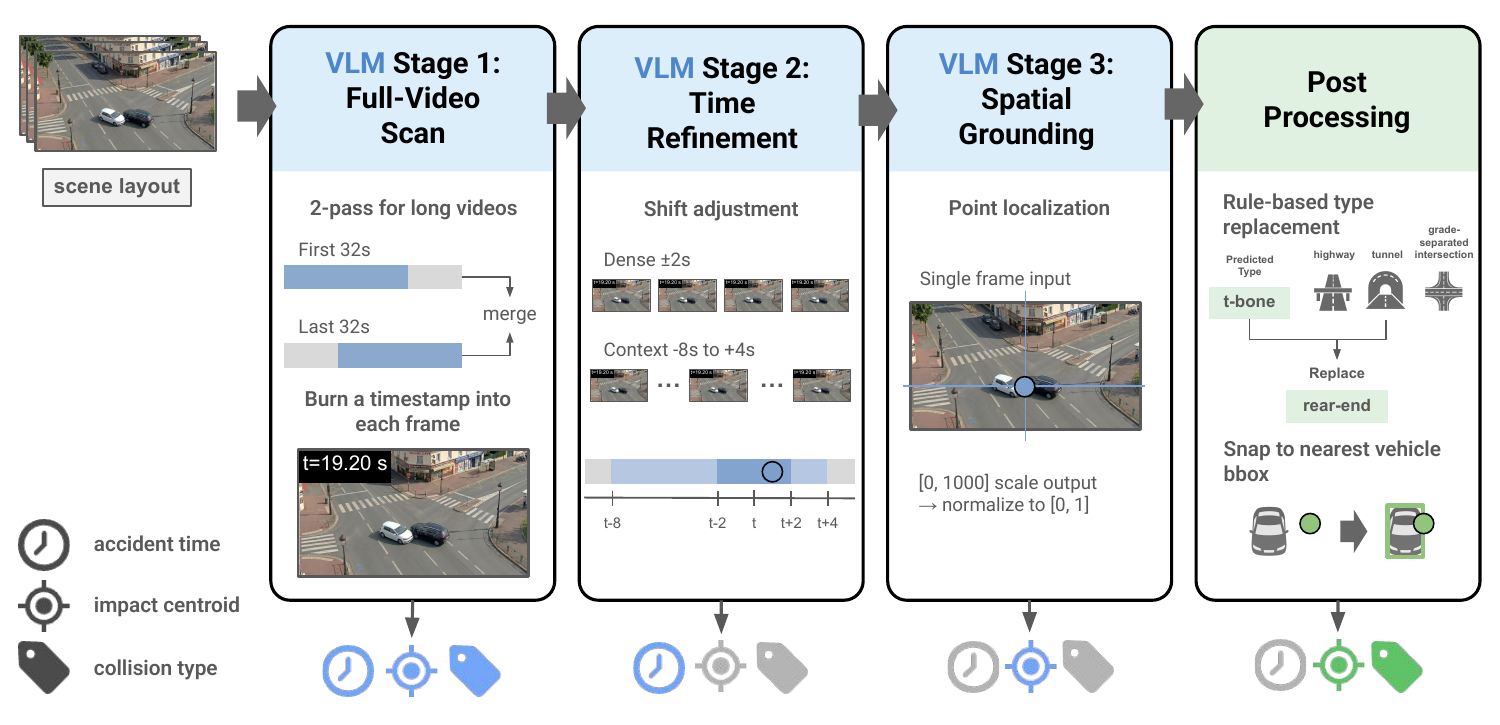}
\caption{Overview of the proposed pipeline.
\textbf{Stage~1}: timestamps are burned into each frame; for long videos the
clip is split into overlapping passes, inferred separately, and merged.
\textbf{Stage~2}: frames around the Stage-1 predicted time $t$ are resampled,
densely near $t$ and sparsely over a wider context window,
to refine the accident time.
\textbf{Stage~3}: a single frame at the refined time is extracted and the
impact centroid is estimated on the model's native $[0,\,1000]$ coordinate
scale, then normalized to $[0,\,1]^2$.
Finally, rule-based type replacement and bounding-box snap (snapping the predicted
point onto the nearest detected vehicle) are applied to produce the final
prediction.}
\label{fig:pipeline}
\end{figure*}

\begin{abstract}
We present the 1st-place solution to the ACCIDENT challenge at the CVPR 2026
AUTOPILOT Workshop, which asks for zero-shot prediction of accident timing,
impact centroid, and collision type from CCTV footage. On a frozen
Qwen3-VL-32B-Instruct checkpoint we build a three-stage pipeline
(full-video joint prediction, time refinement, and single-frame grounding
of the impact centroid),
run the same pipeline a second time on a 235B Mixture-of-Experts
sibling, blend the two outputs 9:1, and finally snap each predicted point
onto the nearest vehicle detection. The final system reaches Public LB 0.55469 / Private LB 0.57080,
roughly $+0.21$ over the strongest host baseline (Molmo-7B, 0.358) and wins
the challenge. We ablate each component, report the negative results that
shaped the final design, and release the code at
\url{https://github.com/fuumin621/cvpr2026-accident-1st-place-solution}.
\end{abstract}

\section{Introduction}
\label{sec:intro}

Automated understanding of Closed Circuit Television (CCTV) accident
footage is essential for rapid incident response and traffic safety
analysis.
The task requires answering three questions from a single clip:
\emph{when} a collision occurred, \emph{where} on the frame the
impact point lies, and \emph{what kind} of collision it was.
The ACCIDENT benchmark~\cite{picek2026accident} jointly evaluates all
three aspects under a \textbf{zero-shot} protocol: no labeled real
video is available for training, and only synthetic clips from
CARLA~\cite{carla} are provided.

In this paper we present the 1st-place solution to the corresponding
ACCIDENT challenge at the CVPR 2026 AUTOPILOT Workshop, built around a three-stage Vision-Language Model
(VLM) pipeline that decomposes the joint prediction into specialized
queries, together with comprehensive ablations and negative results.

Our method makes three successive calls to the same VLM:
\begin{enumerate}
    \item \textbf{Stage~1 --- Full-video scan.}
    The model sees the whole clip with burned-in timestamps and returns
    a first joint estimate of time, location, and type.
    \item \textbf{Stage~2 --- Time refinement.}
    A denser window around the Stage-1 time yields a bounded correction.
    \item \textbf{Stage~3 --- Spatial grounding.}
    A single frame at the refined time is used to pinpoint the
    impact location, replacing the Stage-1 coordinates.
\end{enumerate}
A second run of the same pipeline on a larger Mixture-of-Experts (MoE)
model is blended at low weight, and each predicted point is snapped
onto the nearest vehicle detection (\cref{sec:bbox_snap}).

\paragraph{Contributions.} (i) A three-stage VLM pipeline that splits
accident prediction into specialized queries on a single VLM.
(ii) Ablations and negative results that clarify what helps and what
does not in the zero-shot setting.

\section{ACCIDENT Challenge}
\label{sec:dataset}

\subsection{Task}

Participants are given a set of CCTV clips and must predict, for each
clip, three outputs: (1) the accident time in seconds, (2) the
collision bounding box (bbox) centroid in normalized image coordinates, and
(3) the collision type from a closed set of five classes.
No real labeled clips are provided for training; only synthetic clips
are available (\cref{sec:data}).
Submissions are ranked by the unified metric $ACC^S$ defined in
\cref{sec:metrics}.

\subsection{Dataset}
\label{sec:data}

The ACCIDENT benchmark~\cite{picek2026accident} consists of 2{,}027 real CCTV clips and
2{,}211 CARLA-synthetic clips, each about 30\,s long. A random temporal
offset is applied during trimming so that the collision does not fall
at a fixed position in the clip, which prevents trivial center-bias
solutions.
Every clip is annotated with an accident time, a bbox around
the collision, and a collision type from \emph{head-on, rear-end,
t-bone, sideswipe, single-vehicle}.
Scene metadata (\texttt{scene\_layout}, video quality, weather) is also provided. The ACCIDENT challenge uses the \textbf{zero-shot}
protocol: the real clips are not available for training; only the synthetic
clips are, and their appearance differs substantially from real CCTV. We
observe that this sim-to-real gap is the dominant source of error when
transferring any synthetic-data signal to the test set
(\cref{sec:negative_results}).

\subsection{Evaluation metrics}
\label{sec:metrics}

Each video receives three scores:
\begin{itemize}
    \item \textbf{Temporal (T).} A Gaussian similarity score between predicted and
    ground-truth accident times, averaged over three temporal tolerances
    ($\sigma_t \in \{0.5, 1, 2\}$\,s).
    \item \textbf{Spatial (S).} An anisotropic Gaussian similarity score on the
    predicted impact centroid $(\hat x, \hat y)$, with $(\sigma_x, \sigma_y)$
    set to the mean annotated bbox width and height.
    \item \textbf{Classification (C).} Top-1 accuracy on the 5-class collision
    type.
\end{itemize}

The unified score used for leaderboard ranking is defined as the \textbf{harmonic mean}:
\begin{equation}
ACC^S \;=\; \frac{3}{1/T + 1/S + 1/C} \;\in\; [0, 1].
\end{equation}

The harmonic mean is pulled down by the smallest of $T$, $S$, and $C$:
raising one of them at the cost of another rarely raises $ACC^S$. The
strongest zero-shot baseline in the host paper, Molmo-7B, reaches
end-to-end $T{=}0.343$, $S{=}0.488$, $C{=}0.293$, giving $ACC^S{=}0.358$. Our final
system reaches 0.57080 on the private split.

\section{Method}
\label{sec:method}

\subsection{Overview}
\label{sec:overview}

Our key design principle is to \emph{decompose} the joint prediction
into three specialized VLM calls rather than asking a single query to
solve time, location, and type simultaneously (\cref{fig:pipeline}).
Predicting all three at once forces the model to allocate attention
across competing objectives, which degrades spatial accuracy in
particular.
Therefore, we first obtain a coarse joint estimate from the full clip
(\cref{sec:stage1}), then refine the temporal prediction with a denser
frame window (\cref{sec:stage2}), and finally re-ground the impact
centroid on a single frame at the refined time (\cref{sec:stage3}).
After the three VLM stages, three minor refinements follow:
type post-processing for physically implausible cases (\cref{sec:type_postfix});
a model-scale ensemble that blends a second run from a larger checkpoint (\cref{sec:ensemble});
and a bbox snap step that aligns the predicted point with a detected vehicle (\cref{sec:bbox_snap}).
The primary backbone is Qwen3-VL-32B-Instruct-FP8~\cite{qwen3vl};
the ensemble additionally runs Qwen3-VL-235B-A22B-Instruct (\cref{sec:ensemble}).
Both are served through vLLM~\cite{vllm}, require no training, and are decoded greedily.

\subsection{Stage~1: Full-video scan}
\label{sec:stage1}

Stage~1 takes the full clip and the benchmark-provided
\texttt{scene\_layout} metadata as input.
It returns a single JSON object containing the accident time, the
normalized impact centroid $(x, y) \in [0, 1]^2$, and the collision
type.
This output already constitutes a complete prediction.
We decode the clip at 4\,fps, keep at most 128 frames, and resize each
frame so its longer side is 960\,px.
Clips longer than 32\,s are processed in two overlapping passes and
merged by a rule-based scheme: predictions outside the overlap region
keep the corresponding pass output, while predictions inside the
overlap average the time and coordinates and keep the first pass's type.

Two prompt-side choices matter in practice.
First, we draw a small black label \texttt{t=xx.xx\,s} on every frame
so the VLM can copy the timestamp into its answer rather than estimate
it from visual context.
Second, we pass the \texttt{scene\_layout} value in the prompt so that
the model prefers a collision type that is geometrically possible in
that scene. We refer to this prompt-side choice as the \emph{scene hint}.

\subsection{Stage~2: Time refinement}
\label{sec:stage2}

Stage~2 takes the clip and the Stage-1 time $t_{\text{base}}$ as input,
and returns a refined time $t_{\text{final}}$ that replaces
$t_{\text{base}}$.
The $(x, y)$ and type from Stage~1 are carried forward unchanged.
Because sampling at 4\,fps limits Stage-1 time resolution to about
$\pm 0.25$\,s, we re-query the same VLM on a hybrid frame set.
The set consists of a dense local window around $t_{\text{base}}$
($\pm 2$\,s at 4\,fps, up to 12 frames) and a few sparse context
anchors outside it ($-8$ to $+4$\,s at 0.5\,fps, up to 4 frames),
so that the model can still see the motion leading into and out of
the event.
From this we obtain a new estimate $t_{\text{refined}}$.
Rather than replacing $t_{\text{base}}$ with $t_{\text{refined}}$ directly, we combine them as
\begin{equation}
t_{\text{final}} = t_{\text{base}} + \alpha \cdot
\mathrm{clip}\!\left(t_{\text{refined}} - t_{\text{base}},\; -\delta_{\max},\; +\delta_{\max}\right),
\label{eq:time_refine}
\end{equation}
where $\alpha < 1$ down-weights the correction and $\delta_{\max}$ caps
its magnitude (see \cref{sec:impl} for values).
The correction is thus
downweighted and tightly capped, which limits the damage of a rare
Stage-2 outlier---$ACC^S$ is a harmonic mean and is penalized heavily by
large errors. We verified on public LB that this combination improves
over direct replacement ($\alpha=1$, no clip).

\subsection{Stage~3: Spatial grounding}
\label{sec:stage3}

Stage~3 passes a single frame extracted at $t_{\text{final}}$ to the
same VLM without any timestamp overlay, and returns a refined
impact centroid $(x, y)$ that replaces the Stage-1 coordinates; time and
type are carried forward. The prompt asks the model to output a point on
Qwen3-VL's native $[0, 1000]$
coordinate scale; we normalize to $[0, 1]^2$. Running this as a separate
stage helps because Stage~1 has to pick a time and a location under a
joint query, and the resulting coordinates tend to
cluster on a coarse grid. Once Stage~2
has pinned down the collision time, Stage~3 only has to answer
\emph{where} in a single frame the impact occurred, which largely
resolves this quantization.

\subsection{Type post-processing}
\label{sec:type_postfix}

We apply one rule: if the predicted type is \emph{t-bone} and the
\texttt{scene\_layout} is a highway, a tunnel, or a grade-separated
intersection, where two vehicles cannot meet perpendicularly,
we replace \emph{t-bone} with \emph{rear-end}.
We experimented with broader rules (e.g.\ rewriting all
\emph{single-vehicle} predictions on highways, or re-classifying by
bbox aspect ratio), but none improved the public LB
(see \cref{sec:negative_results}).

\subsection{Model-scale ensemble}
\label{sec:ensemble}

We run the same three-stage pipeline a second time with
\textbf{Qwen3-VL-235B-A22B-Instruct} (MoE, 22\,B active per token) and
combine the two runs by a weighted average of the time and spatial outputs:
\begin{equation}
x_{\text{ens}} = \lambda \cdot x_{\text{32B}} + (1-\lambda) \cdot x_{\text{235B}},
\end{equation}
where $\lambda = 0.9$.
The collision type is kept from the 32B run only, because the 235B model
chose a less accurate type under the same prompt. We swept $\lambda$
on public LB: the 235B run alone scored below the 32B run, but
$\lambda = 0.9$ gave a small, consistent gain on
both time and location (see \cref{tab:main_results}).

\subsection{Bounding box snap post-processing}
\label{sec:bbox_snap}

After the model-scale ensemble, VLM points $(x,y)$ tend to land near the
collision area but not exactly on a vehicle. We therefore snap each
prediction to a nearby vehicle bbox detected by
RetinaNet~\cite{retinanet}. We convert \texttt{accident\_time} to a frame
index and gather bboxes within $\pm 10$ frames, keeping only
vehicle classes (COCO: car, motorcycle, bus, truck); if no vehicle
appears in that window, we double the window width and retry. Then we
pick the box whose center is closest to $(x,y)$ and clamp the point
into that box. To avoid pulling predictions toward unrelated vehicles,
the snap is cancelled whenever the induced 2D displacement exceeds a
threshold $\delta_{\text{snap}}$ (normalized coordinates).

\subsection{Implementation summary}
\label{sec:impl}

Stage~1: 4\,fps, $\le$128 frames at 960\,px longest side, 32\,s pass
length with two-pass split for longer clips.
Stage~2: $\alpha=0.35$, $\delta_{\max}=1.5$\,s.
Ensemble weight: $\lambda=0.9$.
Bbox snap: $\delta_{\text{snap}}=0.2$, frame window $\pm 10$.
No training is performed.
Full configuration files are included in the code release.

\section{Experiments}
\label{sec:experiments}

\subsection{Main results}
\label{sec:main_results}

\cref{tab:main_results} shows final leaderboard scores for our system and
its components. The ensemble with bbox snap reaches \textbf{0.55469 public /
0.57080 private}, our 1st-place submission. The 32B pipeline alone already
exceeds the strongest host baseline by a wide margin; the 235B ensemble
adds roughly $+0.002$ on both splits; bbox snap adds another
$+0.00054{/}+0.00132$ on top.

The 32B leg processes all 2{,}027 test clips in approximately 13\,h on a
single NVIDIA RTX PRO 6000; the 235B leg takes approximately 10\,h on
8$\times$ RTX PRO 6000.

\begin{table}[ht]
\centering\small
\begingroup
\setlength{\tabcolsep}{2.8pt}
\resizebox{0.98\linewidth}{!}{%
\begin{tabular}{@{}>{\raggedright\arraybackslash}p{0.36\linewidth}ccccc@{}}
\toprule
System & ACC$^S$ & C & T & S & Private LB \\
\midrule
Molmo-7B baseline~\cite{picek2026accident} & 0.3580 & 0.2930 & 0.3430 & 0.4880 & - \\
Ours 32B (3-stage) & 0.5637 & 0.5994 & 0.5603 & 0.5350 & 0.56740 \\
Ours 235B (3-stage) & 0.5504 & 0.5703 & 0.5388 & 0.5430 & 0.55670 \\
Ours 9{:}1 ensemble \newline(32B type) & 0.5657 & 0.5994 & 0.5600 & 0.5410 & 0.56948 \\
\textbf{Ours + bbox snap \newline(final)} & \textbf{0.5669} & \textbf{0.5994} & \textbf{0.5600} & \textbf{0.5441} & \textbf{0.57080} \\
\bottomrule
\end{tabular}
}
\endgroup
\caption{Final leaderboard scores and detailed metrics. ACC$^S$ is the unified score, C: classification, T: temporal, S: spatial. Private LB was hidden at submission time.}
\label{tab:main_results}
\end{table}

\subsection{Incremental improvements}
\label{sec:ablation}

\cref{tab:ablation} shows the cumulative public-LB scores along our
overall development path. The largest single gain is Stage~3 pointing
grounding (\cref{sec:stage3}): once Stage~2 has pinned down the
collision time, the VLM returns much better coordinates than under the
Stage-1 joint query over both time and space.
Model scale matters independently of the pipeline structure;
\cref{tab:scale} sweeps the Qwen3-VL family at 512\,px on the
single-call baseline.

\begin{table}[ht]
\centering\small
\resizebox{0.98\linewidth}{!}{%
\begin{tabular}{@{}lrr@{}}
\toprule
Configuration & Public & $\Delta$ \\
\midrule
Single-call VLM (768\,px, 2\,fps)            & 0.42238 & --- \\
\quad + Stage-3 pointing grounding            & 0.51594 & $\mathbf{+0.09356}$ \\
\quad + scene hint in Stage-1 prompt (\cref{sec:stage1})   & 0.51810 & $+0.00216$ \\
\quad + t-bone post-processing                & 0.52276 & $+0.00466$ \\
\quad + Stage-2 time refinement                & 0.53246 & $+0.00970$ \\
\quad + 4\,fps / 128\,f / 960\,px             & 0.55215 & $+0.01969$ \\
\quad + 9:1 ensemble with 235B                & 0.55415 & $+0.00200$ \\
\quad + bbox snap (final)                     & \textbf{0.55469} & $+0.00054$ \\
\bottomrule
\end{tabular}
}
\caption{Cumulative public-LB scores along our overall development
path. \emph{Single-call VLM} (first row) is the baseline in which a
single VLM query jointly predicts accident time, impact centroid, and
collision type from the full clip at 768\,px and 2\,fps, without
Stages~2 or~3.
The first six rows build the 32B pipeline; the last two rows
add the 9:1 ensemble with 235B and the bbox snap post-processing.
Each row adds a single component on top of the previous row; the
largest single $\Delta$ is bold.}
\label{tab:ablation}
\end{table}

\begin{table}[ht]
\centering\small
\begin{tabular}{@{}lc@{}}
\toprule
Backbone (Qwen3-VL, 512\,px) & Public \\
\midrule
4B  & 0.285 \\
8B  & 0.365 \\
32B & 0.387 \\
\bottomrule
\end{tabular}
\caption{Backbone-scale sweep on the single-call VLM baseline at
512\,px (no Stage-2/3, no ensemble).}
\label{tab:scale}
\end{table}

\subsection{Negative results}
\label{sec:negative_results}

The following approaches were submitted or ablated but did not improve the
public LB. We group them by failure mode.

\begin{itemize}\itemsep 2pt
    \item \textbf{Sampling / decoding tricks.} CoT (``thinking'' mode)
    collapsed type accuracy and reduced the temporal score from 0.42 to
    0.06; self-consistency ($3\times$ at temperatures
    0.1--0.6) and flip TTA each changed fewer than $\sim$10 predictions
    on 2{,}027 clips, i.e.\ effectively noise.
    Raising the sampling rate from 4 to 10\,fps also degraded the score
    by $-0.009$; an excessively dense frame rate appears to decrease
    stage-level accuracy rather than improve it.
    \item \textbf{Richer grounding inputs.} Three-frame grounding averaged
    coordinates and decreased the spatial score; rendering the grounding
    frame at 1024 or 1280\,px also degraded spatial accuracy.
    \item \textbf{Visual input enrichment for type classification.}
    Overlaying object-detection tracking trails on input
    frames, which was intended to compress temporal dynamics into a
    single frame and improve type classification, consistently degraded
    performance.
    \item \textbf{Alternative VLMs.} We evaluated Cosmos-Reason2-8B, Qwen3.5-35B-A3B
    (text-focused MoE), InternVL3.5-8B, Gemma~3-27B, and MiniCPM-V~4.5
    as alternative backbones; none matched Qwen3-VL-32B on the combined
    metric. A consistent pattern emerged: non-VL-specialized models
    (Qwen3.5-35B-A3B) achieved competitive temporal scores but collapsed on
    type classification and spatial accuracy; Cosmos-Reason2 missed
    the \emph{t-bone} class entirely; InternVL3.5-8B and Gemma~3-27B
    mapped most predictions to \emph{rear-end}.
    \item \textbf{Synthetic-data transfer.} A CNN classifier trained on
    CARLA, an optical-flow + detector hybrid, and a frame-offset ensemble
    all improved sim validation but decreased the LB score.
    \item \textbf{Over-aggressive post-processing.} Broadening the
    t-bone rewrite beyond the physically-impossible subset degraded the
    leaderboard score.
\end{itemize}

\section{Conclusion}
\label{sec:conclusion}

Our winning entry is a three-stage VLM pipeline on a single
Qwen3-VL-32B-Instruct-FP8 checkpoint, blended 9:1 with the same pipeline
on a 235B MoE sibling and followed by a bbox snap post-processing step.
Three design choices account for most of the score: splitting the joint
prediction into three specialized VLM calls, asking for the impact centroid
on a single frame at the refined time, and applying a small, bounded
correction in the time refinement step rather than replacing the Stage-1
time. The system is training-free, and our reported 32B runs finish in
about 13 hours on a single RTX PRO 6000, reaching 0.57080 on the
private leaderboard---about $+0.21$ over the strongest host baseline.

{
    \small
    \bibliographystyle{ieeenat_fullname}
    \bibliography{refs}
}

\end{document}